\documentclass[]{spie}  %>>> use for US letter paper
\usepackage{amsmath,amsfonts,amssymb}
\usepackage{graphicx}
\usepackage[colorlinks=true, allcolors=blue]{hyperref}
\usepackage{float} % force figures to stay ``here''

\title{Deep convolutional networks for automated detection of posterior-element fractures on spine CT}

\author[a]{Holger R. Roth}
\author[a]{Yinong Wang}
\author[a]{Jianhua Yao}
\author[a]{Le Lu}
\author[b]{Joseph E. Burns}
\author[a]{\\ Ronald M. Summers}
\affil[a]{Imaging Biomarkers and Computer-Aided Diagnosis Laboratory, Radiology and Imaging Sciences, National Institutes of Health Clinical Center, Bethesda, MD 20892-1182, USA.}
\affil[b]{Department of Radiological Sciences, University of California-Irvine, Orange, CA 92868, USA.}

\authorinfo{Further author information contact Holger Roth (\url{holger.roth@nih.gov}) or Ronald Summers (\url{rms@nih.gov})}

\begin{document} 
\maketitle

\begin{abstract}
Injuries of the spine, and its posterior elements in particular, are a common occurrence in trauma patients, with potentially devastating consequences. Computer-aided detection (CADe) could assist in the detection and classification of spine fractures. Furthermore, CAD could help assess the stability and chronicity of fractures, as well as facilitate research into optimization of treatment paradigms. 

In this work, we apply deep convolutional networks (ConvNets) for the automated detection of posterior element fractures of the spine. First, the vertebra bodies of the spine with its posterior elements are segmented in spine CT using multi-atlas label fusion. Then, edge maps of the posterior elements are computed. These edge maps serve as candidate regions for predicting a set of probabilities for fractures along the image edges using ConvNets in a 2.5D fashion (three orthogonal patches in axial, coronal and sagittal planes). We explore three different methods for training the ConvNet using 2.5D patches along the edge maps of `positive', i.e. fractured posterior-elements and `negative', i.e. non-fractured elements.

An experienced radiologist retrospectively marked the location of 55 displaced posterior-element fractures in 18 trauma patients. We randomly split the data into training and testing cases. In testing, we achieve an area-under-the-curve of 0.857. This corresponds to 71\% or 81\% sensitivities at 5 or 10 false-positives per patient, respectively. Analysis of our set of trauma patients demonstrates the feasibility of detecting posterior-element fractures in spine CT images using computer vision techniques such as deep convolutional networks.
\end{abstract}

\keywords{spine CT, computer-aided detection, posterior-element fractures, deep learning}
%%%%%%%%%%%%%%%%%%%%%%%%%%%%%%%%%%%%%%%%%%%%%%%%%%%%%%%
%%%%%%%%%%%%%%%%%%%%%%%%%%%%%%%%%%%%%%%%%%%%%%%%%%%%%%%
%%%%%%%%%%%%%%%%%%%%%%%%%%%%%%%%%%%%%%%%%%%%%%%%%%%%%%%
%%%%%%%%%%%%%%%%%%%%%%%%%%%%%%%%%%%%%%%%%%%%%%%%%%%%%%%
\section{INTRODUCTION}
\begin{figure}[H]
  \begin{center}
   \begin{tabular}{c} 
     \includegraphics[width=0.75\textwidth]{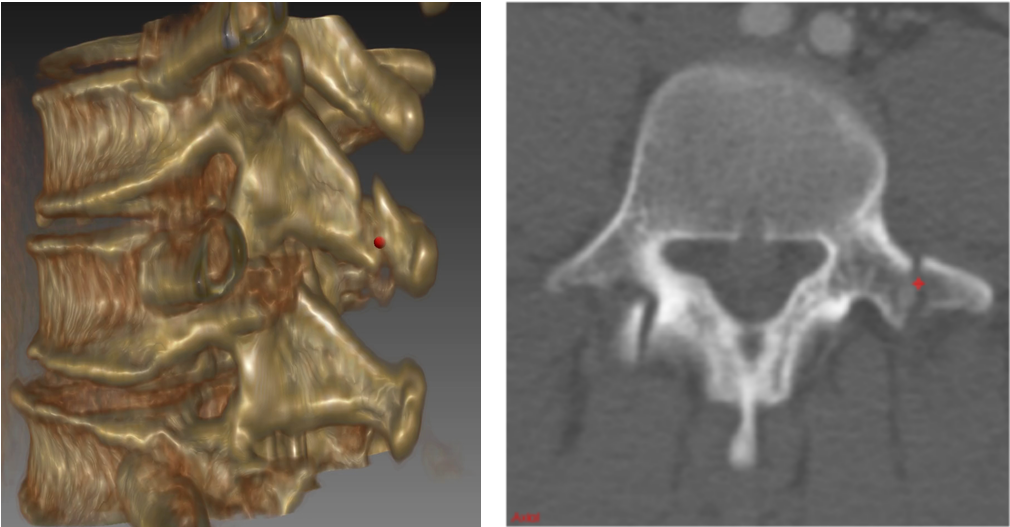}
   \end{tabular}
  \end{center}
  \caption[example] 
  { \label{fig:fig_fracture} 
Example 3D rendering of a posterior element fracture of the spinous process (left) and reformatted images (right) show a posterior element fracture of the transverse process in axial and sagittal views. Manual gold standard marks by radiologist are shown in red.}
\end{figure}

Injuries of the spine and its posterior elements in particular (see Figure \ref{fig:fig_fracture}) are a common occurrence in traumatic patients, with potentially devastating consequences \cite{parizel2010trauma}. Spine fractures are detected using volumetric imaging such as computed tomography (CT) in order to access the degree of injury. Spine injuries are a critical concern in blunt trauma, particularly in cases of motor vehicle collision and fall from significant heights. More than 140,000 vertebral fractures occur in the U.S. each year \cite{yao2014cortical}. However, the traditional method of qualitative visual assessment of images for diagnosis could miss fractures, and is time-consuming, potentially causing delays in time-critical situations such as the treatment of spine injuries. Computer-aided detection (CADe) has the potential to expedite the assessment of trauma cases, reduce the chance of misclassification of fractures of the spine, and decrease inter-observer variability. Furthermore, CADe could help assess the stability and chronicity of fractures, as well as facilitate research into optimization of treatment paradigms.
%%%%%%%%%%%%%%%%%%%%%%%%%%%%%%%%%%%%%%%%%%%%%%%%%%%%%%%
%%%%%%%%%%%%%%%%%%%%%%%%%%%%%%%%%%%%%%%%%%%%%%%%%%%%%%%
%%%%%%%%%%%%%%%%%%%%%%%%%%%%%%%%%%%%%%%%%%%%%%%%%%%%%%%
\section{METHODS}
\label{sec:methods}
Recently, the availability of large annotated training sets and the accessibility of affordable parallel computing resources via GPUs has made it feasible to train deep convolutional networks (ConvNets), also popularized under the keyword “deep learning”, for computer vision classification tasks. Great advances in classification of natural images have been achieved \cite{krizhevsky2012imagenet,zeiler2014visualizing}. Studies that have tried to apply deep learning and ConvNets to medical imaging applications also showed promise \cite{yan2015bodypart,cirecsan2013mitosis,roth2014new,roth2015improving}. 
%%%%%%%%%%%%%%%%%%%%%%%%%%%%%%%%%%%%%%%%%%%%%%%%%%%%%%%
\subsection{Convolutional networks}
\begin{figure}[H]
  \begin{center}
   \begin{tabular}{c} 
     \includegraphics[width=0.75\textwidth]{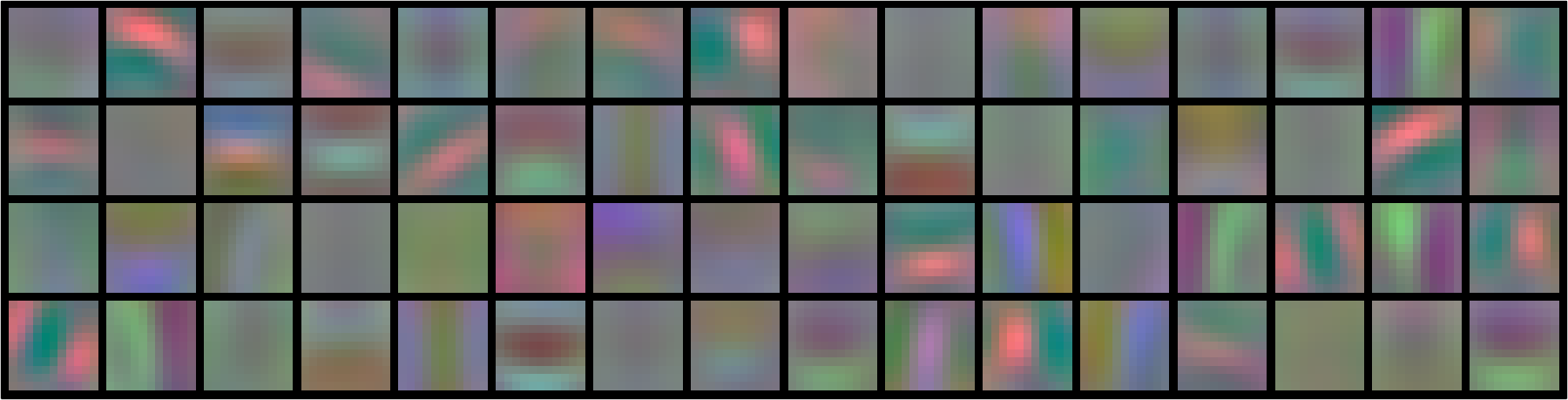}
   \end{tabular}
  \end{center}
  \caption[example] 
  { \label{fig:conv_filters} 
	Convolution filters trained on examples of posterior-element fractures and non-fractures.
}
\end{figure}

ConvNets are named for their convolutional filters which are used to compute image features for classification \cite{krizhevsky2012imagenet}. In this work, we use a ConvNet similar to \cite{krizhevsky2012imagenet}\footnote{\url{https://code.google.com/archive/p/cuda-convnet2/}} with 5 convolutional layers, 3 fully connected layers, and a final softmax layer for classificaton. However, we use input images of size $64 \times 64$ and a stride of 1 in the first convolutional layer. All convolutional filter kernel elements are trained from the data in a fully-supervised fashion. This has major advantages over more traditional CADe approaches that use hand-crafted features, designed from human experience. ConvNets have a better chance of capturing the ``essence'' of the imaging data set used for training than when using hand-crafted features \cite{krizhevsky2012imagenet,zeiler2014visualizing}. Furthermore, we can train the similarly configured ConvNet architectures from randomly initialized or pre-trained model parameters for detecting different lesions or pathologies (with heterogeneous appearances), with no manual intervention of system and feature design. Examples of trained filters from the first convolutional layer for posterior-element fracture detection are shown in Fig. \ref{fig:conv_filters}. Dropout \cite{srivastava2014dropout} is used during training as a form of regularization that avoids overfitting to the training data. 
%%%%%%%%%%%%%%%%%%%%%%%%%%%%%%%%%%%%%%%%%%%%%%%%%%%%%%%
%%%%%%%%%%%%%%%%%%%%%%%%%%%%%%%%%%%%%%%%%%%%%%%%%%%%%%%
\subsection{Application to spine CT}
\begin{figure}[H]
  \begin{center}
   \begin{tabular}{c} 
     \includegraphics[width=0.95\textwidth]{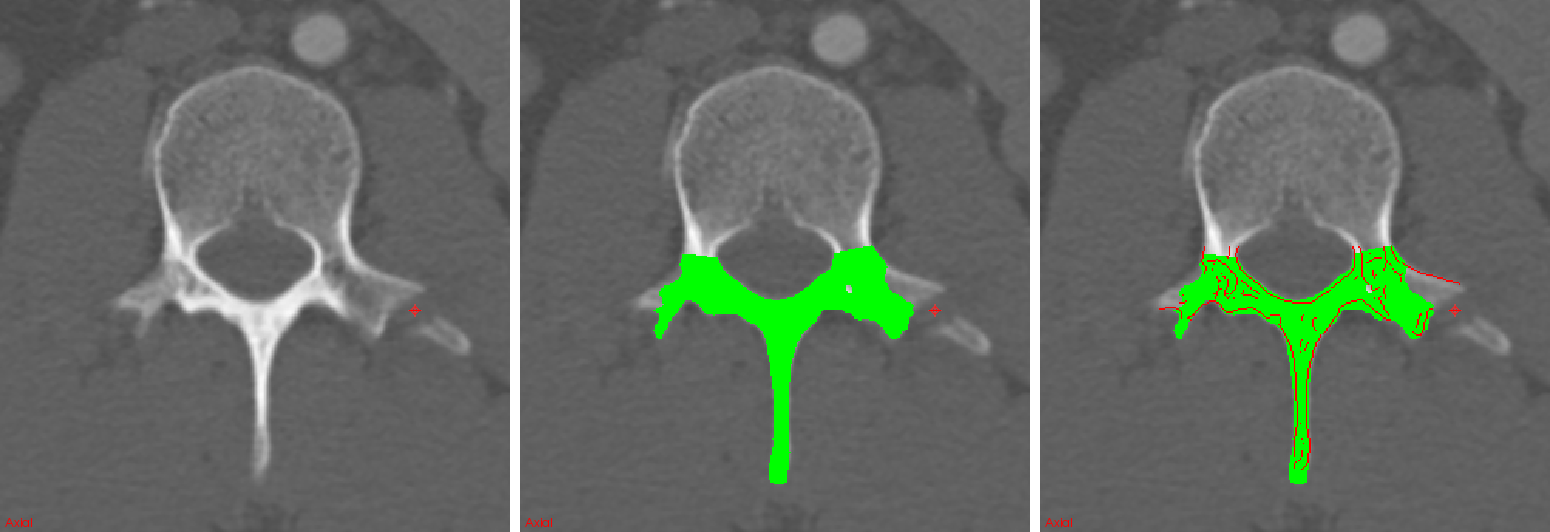}
   \end{tabular}
  \end{center}
  \caption[example] 
  { \label{fig:candidate_post_elements} 
(a) Posterior element fracture; (b) posterior element segmentation using multi-atlas label fusion\cite{ywang2015multi}; (c) edge map estimation.
}
\end{figure}

\begin{figure}[H]
  \begin{center}
   \begin{tabular}{c} 
     \includegraphics[width=0.95\textwidth]{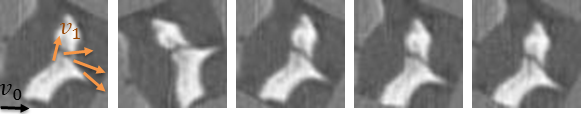}
   \end{tabular}
  \end{center}
  \caption[example] 
  { \label{fig:oriented} 
Illustration of 2D patches aligned along the candidate edge orientation  with respect to the origin  for training convolutional networks (ConvNets).
}
\end{figure}

In this work, we apply ConvNets for the automated detection of posterior element fractures of the spine. First, the vertebra bodies of the spine with its posterior elements are segmented in spine CT using multi-atlas label fusion \cite{ywang2015multi,wang2013multi}. A set of atlases of the vertebra bodies are registered to the target image with free-form deformation. Then, an edge map $E$ of the posterior elements is computed using the Sobel operators
\begin{equation}
S_x = \begin{Bmatrix} -1 & 0 & +1 \\ -2 & 0 & +2 \\ -1 & 0 & +1 \end{Bmatrix} \ \ \mathrm{and} \ \ S_y = \begin{Bmatrix} -1 & -2 & -1 \\ 0 & 0 & 0 \\ +1 & +2 & +1 \end{Bmatrix}
\end{equation}
in order to find the horizontal ($G_x = S_x*I$) and vertical ($G_y = S_x*I$) approximations of the image derivative for each CT slice. Here, $*$ denotes a convolutional operation. Edge points are located at the maximum of the absolute gradient $\left|G\right| = \sqrt{G_x^2 + G_y^2}$. The edge maps $E$ serve as candidate regions for predicting a set of probabilities $P$ for fractures along an image edge using ConvNets. An example of posterior-element segmentation and edge map estimation is shown in Fig. \ref{fig:candidate_post_elements}.
%%%%%%%%%%%%%%%%%%%%%%%%%%%%%%%%%%%%%%%%%%%%%%%%%%%%%%%
%%%%%%%%%%%%%%%%%%%%%%%%%%%%%%%%%%%%%%%%%%%%%%%%%%%%%%%
\clearpage
\newpage
We explore three different methods for training the ConvNet using 2.5D patches (three orthogonal patches in axial, coronal and sagittal planes) along the edge maps of ‘positive’, i.e. fractured posterior-elements and ‘negative’, non-fractured elements:
%%%%%%%%%%%%%%%%%%%%%%%%%%%%%%%%%%%%%%%%%%%%%%%%%%%%%%%
\begin{enumerate}
  \item $P_\mathrm{original}$: 2.5D patches along each edge voxel $e \in E$ are aligned to original scanner coordinates in axial, coronal, and sagittal planes.
  \item $P_\mathrm{mirrored}$: 2.5D patches at voxels $e$ near fractures are mirrored along the $y$-axis of the image in order to increase ‘positive’ training examples.
  \item $P_\mathrm{oriented}$: 2.5D patches are oriented along the principal axis of an edge at voxel $e$ (while ‘positive’ training examples are handled as in case 2, i.e. they are also mirrored).
\end{enumerate}
%%%%%%%%%%%%%%%%%%%%%%%%%%%%%%%%%%%%%%%%%%%%%%%%%%%%%%%
The alignment of a training patch along its edge orientation is illustrated in Fig. \ref{fig:oriented}. This step can be also seen as a form of data augmentation that artificially increases the variation of training examples (as opposed to just having slightly shifted versions of nearby patches as in case 1 and 2). Orientating a patch along the anatomical structures of interest has also been shown to improve performance in other applications such as in detection of pulmonary embolism \cite{tajbakhsh2015computer} where the image patch can be aligned along the direction of a blood vessel. The orientation of an edge is estimated as in Equation \ref{equ:orientation}. In 2D, the eigenvector  $v_1$ corresponding to the largest eigenvalue $\lambda_1$ corresponds to the major axis of a local edge. Hence, the orientation $\theta$ of the edge can be estimated as the angle between this eigenvector and the origin $v_0$:
%%%%%%%%%%%%%%%%%%%%%%%%%%%%%%%%%%%%%%%%%%%%%%%%%%%%%%%
\begin{equation}
	\theta = \tan^{-1}\left(\frac{v_1}{v_0}\right)
	\label{equ:orientation}
\end{equation}
%%%%%%%%%%%%%%%%%%%%%%%%%%%%%%%%%%%%%%%%%%%%%%%%%%%%%%%
%%%%%%%%%%%%%%%%%%%%%%%%%%%%%%%%%%%%%%%%%%%%%%%%%%%%%%%
%%%%%%%%%%%%%%%%%%%%%%%%%%%%%%%%%%%%%%%%%%%%%%%%%%%%%%%
\section{RESULTS}
\subsection{Data}
\begin{figure}[H]
  \begin{center}
   \begin{tabular}{c} 
     \includegraphics[width=0.95\textwidth]{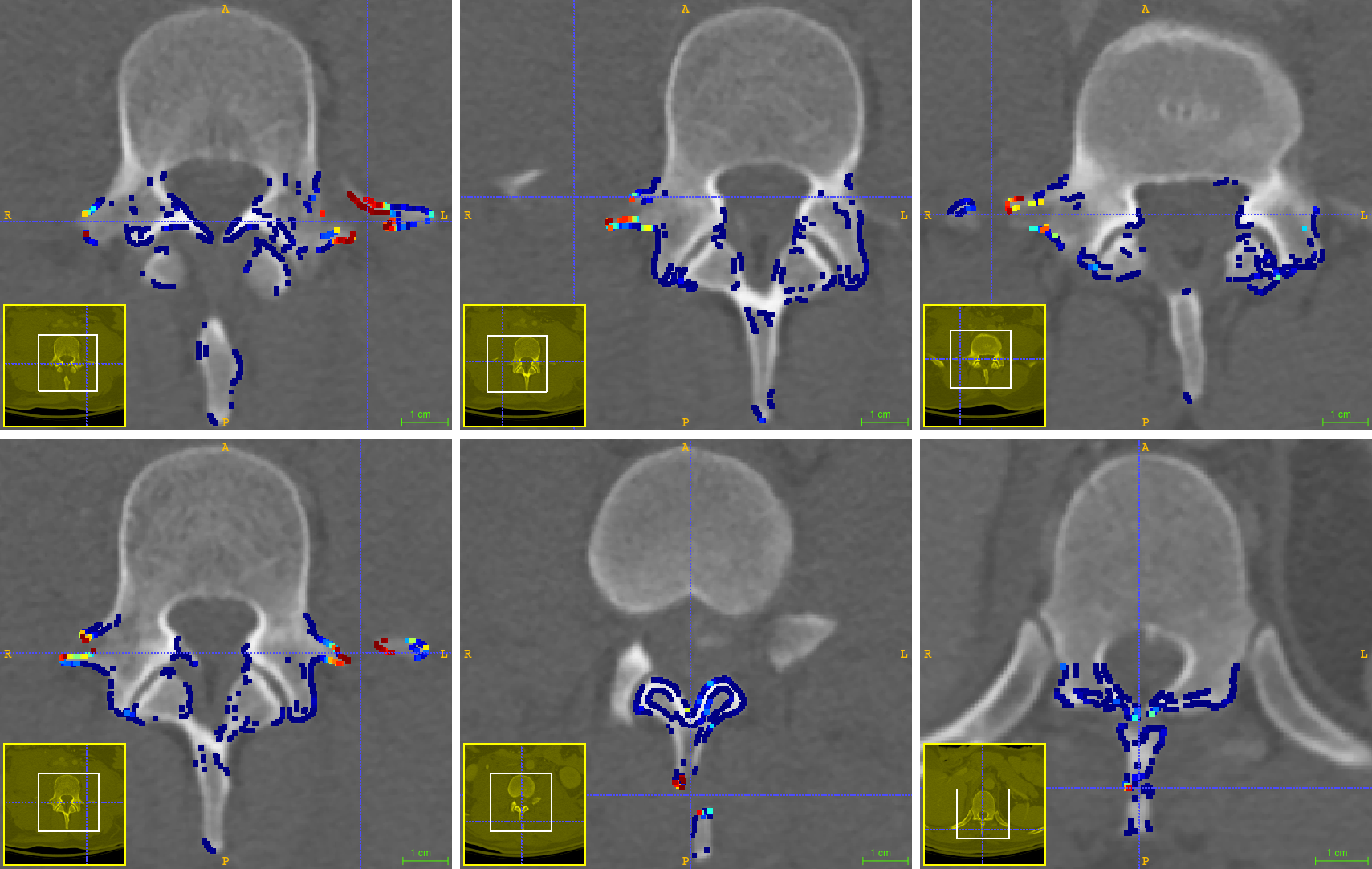}
   \end{tabular}
  \end{center}
  \caption[example] 
  { \label{fig:prob_maps} 
Examples of probability maps for posterior-element fracture detection along candidate edges. The posterior element fractures are marked by crosshairs.
}
\end{figure}

An experienced radiologist retrospectively marked the location of 55 displaced posterior-element fractures in 18 trauma patients admitted for traumatic emergencies to the University of California-Irvine Medical Center. Image sizes range within $512\times 512\times [259-582]$ and resolutions range within $[0.29-0.41]\times [0.29-0.41]\times [1.0-2.0]$ mm. We use a random subset of 12 of these patients with spine fractures for training ConvNets as described in Section \ref{sec:methods}; 6 patients are reserved for testing. An additional set of 5 spine CTs of healthy patients were added to the training set in order to increase the number of non-fractured examples. A total of ~800,000 2.5D patches are randomly selected from the candidate edge maps $\left\{E\right\}$ of the training set and used for learning the ConvNet parameters. After convergence, the ConvNet is applied to a testing case edge map E in order to produce a probability map $P$ for fractures. Figure \ref{fig:prob_maps} shows examples of probability maps for posterior-element fracture detection along candidate edges. The radiologist's markings are indicated by crosshairs.
%%%%%%%%%%%%%%%%%%%%%%%%%%%%%%%%%%%%%%%%%%%%%%%%%%%%%%%
%%%%%%%%%%%%%%%%%%%%%%%%%%%%%%%%%%%%%%%%%%%%%%%%%%%%%%%
\subsection{Performance}
\begin{figure}[H]
  \begin{center}
   \begin{tabular}{c} 
     \includegraphics[width=0.95\textwidth]{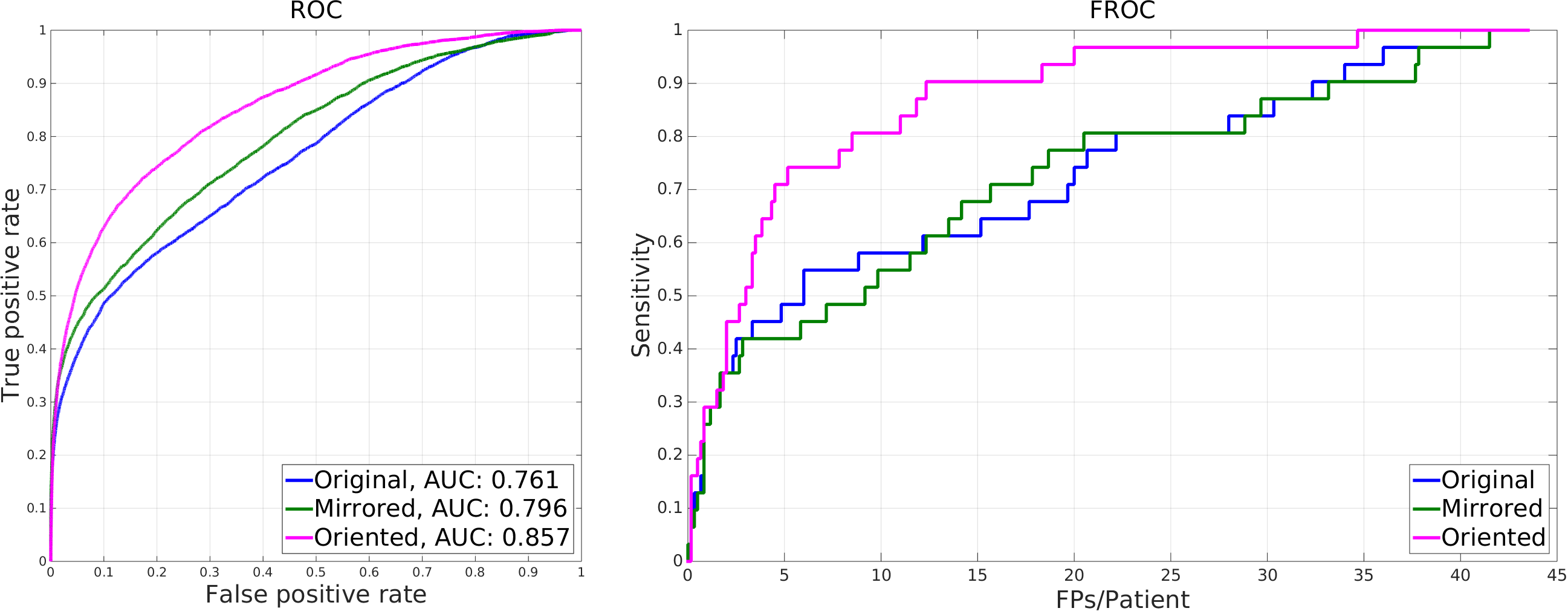}
   \end{tabular}
  \end{center}
  \caption[example] 
  { \label{fig:froc} 
Voxelwise-ROC performance (left) and per-fracture FROC performance (right) evaluating the classification accuracy.
}
\end{figure}

The classification performance is evaluated using voxel-wise ROC and per posterior-element process (left, right and spinous process) FROC analysis. Fig. \ref{fig:froc} shows the voxel-wise ROC and processes-wise FROC performance using the differently trained ConvNets. A clear advantage of using `Oriented' patches for training can be observed in testing, compared to `Original' and `Mirrored' patches for training. In testing, we achieve area-under-the-curve (AUC) values of 0.761, 0.796 and 0.857 for `Original', `Mirrored', and `Oriented' patch training respectively. This corresponds to 71\% or 81\% sensitivities at 5 or 10 false-positives per patient, respectively, in the case of training and testing with `Oriented' patches.
%%%%%%%%%%%%%%%%%%%%%%%%%%%%%%%%%%%%%%%%%%%%%%%%%%%%%%%
%%%%%%%%%%%%%%%%%%%%%%%%%%%%%%%%%%%%%%%%%%%%%%%%%%%%%%%
%%%%%%%%%%%%%%%%%%%%%%%%%%%%%%%%%%%%%%%%%%%%%%%%%%%%%%%
\section{CONCLUSION}
While there has been work on automated detection of other spinal injuries, such as vertebral compression fractures \cite{yao2014cortical}, to the best of our knowledge, this is the first work to explore automated posterior-element fracture detection. It demonstrates that deep convolutional networks (ConvNets) can be useful for detection tasks, such as the detection of fractures in spine CT. 

Results from analysis of our set of trauma patients demonstrate the feasibility of detecting posterior-element fractures in spine CT images using computer vision techniques such as ConvNets. We evaluated three methods for training a ConvNet and show the advantages of applying an oriented patch extraction method ($P_\mathrm{oriented}$) for better classification performance. 

\acknowledgments % equivalent to \section*{ACKNOWLEDGMENTS}       
This work was supported by the Intramural Research Program of the NIH Clinical Center. 
%%%%%%%%%%%%%%%%%%%%%%%%%%%%%%%%%%%%%%%%%%%%%%%%%%%%%%%
%%%%%%%%%%%%%%%%%%%%%%%%%%%%%%%%%%%%%%%%%%%%%%%%%%%%%%%
% References
\clearpage 
\newpage
\bibliography{SPIE2016} % bibliography data
\bibliographystyle{spiebib} % makes bibtex use spiebib.bst

\end{document}